\documentclass[runningheads]{llncs}
\usepackage[T1]{fontenc}
\usepackage{amsmath}
\usepackage{amssymb}
\usepackage{amsfonts}
\usepackage{graphicx,verbatim}
\usepackage{booktabs}
\usepackage{arydshln}
\usepackage{caption}    
\usepackage{adjustbox}

\begin{document}
\title{Leveraging Cell-level Spatial Transcriptomics \\as Weak Supervision for Nuclei Classification}
\titlerunning{Cell-level ST as Weak Supervision for Nuclei Classification}
%
\author{Kazuya Nishimura\inst{1} \and
Ryoma Bise\inst{2}\and
Haruka Hirose\inst{3} \and
Yasuhiro Kojima\inst{3}
}
\authorrunning{K. Nishimura et al.}
%
\institute{The University of Osaka, Osaka, Japan \\
\email{k.nishimura.d3c@osaka-u.ac.jp}
\url{https://is.d3c.osaka-u.ac.jp/en/}
\and
Kyushu University, Fukuoka, Japan\\ 
\and 
National Cancer Center Japan, Tokyo, Japan\\
}


  
\maketitle              
\begin{abstract}
Deep learning–based nuclei analysis in histopathology relies heavily on large-scale, expert-curated datasets that are costly and difficult to scale across organs. To address this, we propose a scalable framework leveraging cell-level spatial transcriptomics (ST) as an alternative supervision source for nuclei classification. Cell-level ST provides both gene expression profiles and nucleus masks, enabling automatic generation of image-label pairs without manual annotation. We convert gene expression into cell-type labels and introduce an image-oriented classification strategy to effectively transfer these identities to histopathology images. Evaluated on unseen tissues, our framework achieves comparable segmentation and classification performance compared with conventional supervised approaches, despite using fewer manually annotated datasets. Code is publicly available at \url{https://github.com/naivete5656/CDGE}.
\end{abstract}

\section{Introduction}
Automated nuclei segmentation and classification from pathology images are fundamental tasks in computational pathology and quantitative precision medicine. Despite the remarkable progress of deep neural networks \cite{horst2024cellvit,graham2019hover,cheng2008segmentation,veta2013automatic}, their performance critically depends on large-scale pixel-level annotations. In histopathology, generating such annotations is labor-intensive, costly, and subject to substantial inter-observer variability, thereby limiting scalability across diverse organs and clinical settings \cite{gamper2020pannuke}. Reducing reliance on manually curated pixel-level dataset remains a central challenge.

To address this limitation, we propose a Cross-modal automatic supervision task, in which cell-level gene expression profiles obtained from spatial transcriptomics (ST) serve as supervisory signals for nuclei segmentation and classification. Emerging ST platforms, such as Xenium \cite{janesick2023high}, simultaneously acquire H\&E-stained images, DAPI-stained nuclear images, and spatially resolved gene expression measurements from the same tissue section.

Although generating new cell-level ST data remains costly, paired image--molecular datasets are increasingly released and aggregated in reusable public resources. Driven by strict research data policies from publishers \cite{nature_reporting_standards,springernature_research_data_policy} and massive tissue-mapping consortia such as the Human BioMolecular Atlas Program (HuBMAP) \cite{hubmap_consortium_2019,hubmap_data_portal}, a vast amount of multimodal ST data is continuously deposited into public domains and curated bio-repositories \cite{jaume2024hest,barrett2013ncbi}.
However, there is currently no established framework for deriving reliable image-level classification labels directly from ST data.

We propose a practical training framework that leverages ST data for pathology image analysis. DAPI-derived nuclear masks are used as segmentation labels, while class labels are inferred from gene expression profiles. Since transcriptomic taxonomy is often finer-grained than morphology-based categories, we define image-level classes according to clinically established pathology classifications. To handle ambiguity arising from measurement noise and the continuous nature of gene expression, we introduce an explicit \textit{Unknown} category for cases where confident assignment is not feasible.
Following pathological classification taxonomy, we adopt a two-stage classification strategy that explicitly models neoplastic identification.

We first evaluate segmentation robustness by assessing generalization to previously unseen organs and comparing with models trained on manually annotated datasets. We then evaluate classification performance and compare our labeling strategy with existing gene expression–based cell-typing approaches, demonstrating the effectiveness of the proposed task formulation.

The main contributions of this study are as follows:

\begin{itemize}
\item We formulate nuclear segmentation and classification using spatial transcriptomics as a cross-modal weakly supervised (cross-modal automatic) learning problem, and position ST as a reusable, potentially scalable supervision source.
\item We propose a practical label generation framework with a hierarchical design and an explicit \textit{Unknown} category to mitigate label noise.
\item We demonstrate improved segmentation generalization to unseen organs and superior classification performance compared with existing transcriptomics-based labeling strategies.
\end{itemize}

\section{Related Work}

\noindent
{\bf Cell Nuclei Segmentation.}
Nuclei segmentation and classification in H\&E-stained histopathology images is a core task in computational pathology. Early methods relied on handcrafted processing techniques \cite{cheng2008segmentation,veta2013automatic}. With the availability of large annotated datasets \cite{kumar2019multi,gamper2019pannuke,graham2021lizard}, deep learning has become the dominant paradigm. Instance-aware models such as HoVer-Net~\cite{graham2019hover} and CellViT~\cite{horst2024cellvit} address overlapping nuclei by jointly modeling segmentation and structural cues.

However, these approaches require extensive pixel-level expert-curated dataset. In contrast, we leverage spatial transcriptomics (ST) as weak supervision.

\noindent
{\bf Nuclei Segmentation with Spatial Transcriptomics.}
Recent work has explored segmentation methods tailored to ST data. SCS~\cite{chen2023scs} and RNA2seg~\cite{defard2025rna2seg} learn segmentation directly from gene expression signals. Multimodal approaches, such as GeneSegNet~\cite{wang2023genesegnet} and BIDCell~\cite{petukhov2022cell}, integrate gene expression with histological images to improve segmentation accuracy.

While effective, these methods require gene expression measurements at inference time, which may limit generalizability across platforms. 

\noindent
{\bf Cell-Type Annotation in Spatial Transcriptomics.}
Cell-type annotation in ST largely builds on methods developed for single-cell RNA sequencing \cite{cheng2025benchmarking}. Marker-based approaches assign cell identities using predefined gene signatures \cite{aibar2017scenic,andreatta2021ucell}, but their performance depends on the robustness of selected markers. 
Reference-based methods, including cell2location \cite{kleshchevnikov2022cell2location}, RCTD \cite{cable2022robust}, and SPOTlight \cite{elosua2021spotlight}, leverage scRNA-seq data to infer cell-type composition in ST data. 
These approaches require appropriate reference datasets, whose availability and transferability may be limited across tissues and disease conditions. 

Importantly, these methods are primarily designed for molecular analysis and are not designed for image classification.

\section{Cross-modal Weak Supervised Learning}
We formulate nuclear segmentation and classification in histopathology as a cross-modal weakly supervised learning problem, where spatial transcriptomics (ST) provides supervisory signals without manual pixel-level annotation.

Let $(S, \{ (b_j, g_j) \}_{j=1}^{N_{\mathrm{cell}}})$ denote the data acquired using a Xenium platform~\cite{janesick2023high}, where $S$ is a high-resolution whole-slide image (WSI) obtained from tissue type $t$, $b_j$ is nuclear boundary mask derived from DAPI staining, and $g_j$ is the corresponding gene expression profile with $g_j \in \mathbb{R}^{N_g}$.

Since transcriptomic measurements provide molecular characterization rather than image-level class labels, we define a mapping
$\phi: \mathbb{R}^{N_g} \rightarrow \mathcal{Y}$, that converts each gene expression vector $g_j$ into a discrete cell-type label $y_j \in \mathcal{Y}$. This process yields cell-level labels $\{ y_j \}_{j=1}^{N_{\mathrm{cell}}}$ inferred from molecular information rather than manual image annotation.

To enable model training, we partition the WSI $S$ into image patches $\{ x_i \}_{i=1}^{N_{\mathrm{patch}}}$. 
For each patch $x_i$, we define the index set $\mathcal{J}_i = \{ j \mid b_j \cap x_i \neq \emptyset \}$,
which contains the nuclei located within the spatial extent of $x_i$.
The resulting training dataset is given by $\mathcal{D} = 
\left\{
x_i,\; t,\;
\left\{ (b_j, y_j) \right\}_{j \in \mathcal{J}_i}
\right\}_{i=1}^{N_{\mathrm{patch}}},
$
where each patch is associated with the tissue type $t$, nuclear segmentation masks $\{ b_j \}_{j \in \mathcal{J}_i}$, transcriptomics-derived cell-type labels $\{ y_j \}_{j \in \mathcal{J}_i}$.
Given $\mathcal{D}$, the objective is to train a deep neural network 
$f_\theta: x_i \mapsto \left( \hat{M}_i, \hat{Y}_i \right)$, that jointly predicts a nuclear segmentation map $\hat{M}_i$, a set of nucleus-level class predictions $\hat{Y}_i$.

The supervision is weak in the sense that segmentation masks and class labels are not manually annotated from histopathology images, but instead derived automatically from ST measurements. 

\begin{figure}[t]
    \centering
    \includegraphics[width=\linewidth]{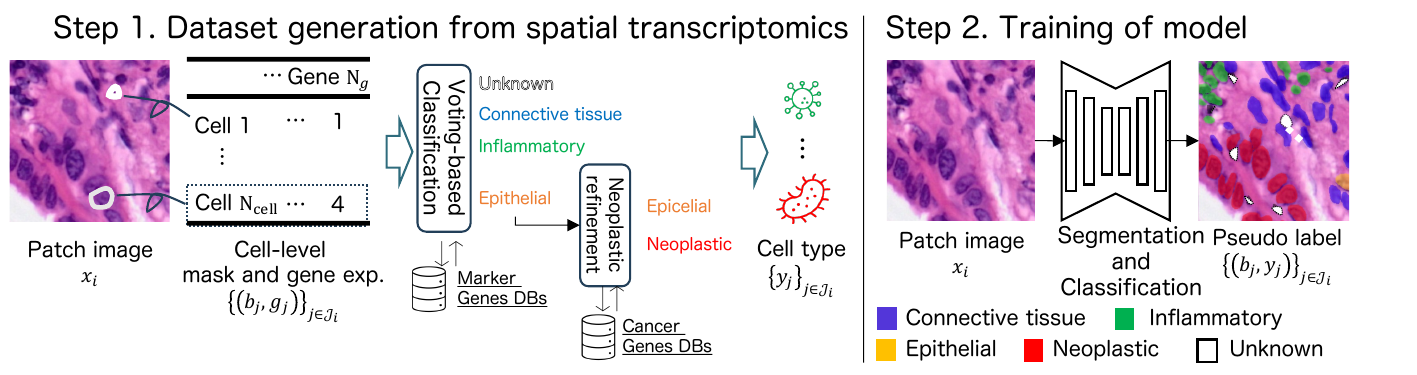}    
    \caption{Overview of proposed framework.}
    \label{fig:overview}
\end{figure}

\section{Spatial Transcriptomics–Guided Model Training}
An overview of the proposed framework is shown in Fig.~\ref{fig:overview}.
Following the formulation above, the pipeline consists of two steps:
1) dataset generation from ST observations, and
2) training a joint segmentation and classification model.
Instead of manually curated cell-type labels, pseudo-labels are derived in three stages: gene expression clustering, marker-based cluster classification, and neoplastic refinement.

In this study, we design the classification taxonomy with the objective of enabling multi-organ recognition based on Pannuke dataset \cite{gamper2019pannuke}.
Although not covered in this study, alternative classification taxonomy can also be adopted depending on the target application, such as organ-specific subtype classification designed by Lizard \cite{graham2021lizard} or PUMA \cite{schuiveling2025novel} datasets.

Unlike PanNuke, which defines a \textit{Dead} category, we alternatively introduce an \textit{Unknown} label in our classification scheme. 
This class is intended to include cells with unstable or unreliable observations, as well as samples for which confident assignment to a specific category is difficult. 

This design is motivated by the nature of gene expression measurements: transcriptomic profiles form a continuous spectrum rather than strictly separable clusters, and the reliability of expression measurements is not uniform across all cells. 
By incorporating an explicit \textit{Unknown} category, we aim to account for ambiguity and variability inherent in the observed molecular data.

\subsection{Dataset Generation from Spatial Transcriptomics}
Based on the defined classes, we adopt a hierarchical classification strategy following the PanNuke taxonomy. 

\noindent
{\bf Transcriptomic Clustering.}
For each ST sample, cells are clustered based on their gene expression profiles using the Leiden algorithm~\cite{traag2019louvain}, applied to a nearest-neighbor graph constructed in principal component space. The resulting cluster assignment $c_j \in \{1, \ldots, K\}$ is used as a proxy for cell identity.

\noindent
{\bf Marker Gene-Based Cluster Classification.}
For each cluster $k$, we compute differentially expressed genes (DEGs) using the Wilcoxon rank-sum test. The top-$N$ ranked genes $\{g_1^k, \ldots, g_N^k\}$ are then matched against two marker gene databases: PanglaoDB~\cite{franzen2019panglaodb}, CellMarker~\cite{hu2023cellmarker}. These databases provide mappings from individual genes to known cell types, which are further mapped to four coarse categories,
$\mathcal{C} = \{\text{Epithelial},\ \text{Inflammatory},\ \text{Connective},\ \text{Unknown}\}$, based on name matching.

For each gene $g_l^k$ at rank $l$, a category vote vector $\mathbf{v}(g_l^k) \in \mathbb{R}^{|\mathcal{C}|}$ is retrieved from the databases. The cluster-level category score is computed as $\mathbf{s}^k = \sum_{l=1}^{N} (M - l)\, \mathbf{v}(g_l^k)$, where the weight $(M - l)$ implements rank-based down-weighting, giving higher-ranked DEGs greater influence. The predicted category for cluster $k$ is $\hat{c}^k = \arg\max_{c \in \mathcal{C}} s_c^k$. The final cell-level label $y_j$ is determined by propagating the cluster-level prediction to all cells belonging to that cluster. 

If total votes $s_c^k$ fall below $\tau_{\text{vote}}$, fallback classification uses the full database. Clusters with insufficient marker evidence are labeled \textit{Unknown} and treated as unlabeled during training.

\noindent
{\bf Neoplastic Refinement.}
Clusters initially classified as \textit{Epithelial} undergo a secondary evaluation to distinguish normal epithelial cells from neoplastic (cancer) cells. We construct two neoplastic-related gene sets from CellMarker and CancerSEA~\cite{yuan2019cancersea}. The fraction of the top-$M$ DEGs overlapping with a curated set of cancer-associated genes is computed. If this ratio exceeds a threshold $\tau_{\text{cancer}}$, the cluster is re-labeled as \textit{Neoplastic}.

Unlabeled or unclassified cells are assigned the Unknown label and are excluded from the supervised loss during training.

\subsection{Training of Nuclei Segmentation and Classification Model}

Given a patch image $x_i$, its nuclear masks and estimated cell-type, and the tissue label $t$, denoted as $\mathcal{Y}_i = \{(b_j, y_j)\}_{j \in \mathcal{J}_i}$, we train a nuclei segmentation and classification model.
We use CellViT \cite{horst2024cellvit}. It has an encoder–decoder structure with three branches for nuclei-related predictions: (1) nuclei probability (binary segmentation), (2) horizontal and vertical distance maps, and (3) nuclei type classification, as well as one additional branch for tissue classification.
For training, we convert the $\mathcal{Y}_i$ into a binary nuclei mask $B \in \mathbb{R}^{H \times W}$, horizontal and vertical distance maps $D \in \mathbb{R}^{H \times W \times 2}$, and a nuclei type mask $S \in \mathbb{R}^{H \times W \times C}$.
\begin{align}
L_{\mathrm{total}}(x_i, \mathcal{Y}_i, t)
&= L_{\mathrm{NP}}(x_i, B)
+ L_{\mathrm{HV}}(x_i, D)
+ L_{\mathrm{NT}}(x_i, S)
+ L_{\mathrm{TC}}(x_i, t),
\end{align}
where $L_{\mathrm{NP}}$, $L_{\mathrm{HV}}$, $L_{\mathrm{NT}}$, and $L_{\mathrm{TC}}$ denote the loss functions for each branches.
For the nuclei type classification loss $L_{\mathrm{NT}}$, pixels annotated as Unknown are excluded from the loss computation.

\section{Experiments}

\noindent
{\bf Implementation details.}
To ensure a fair comparison and reproducibility, all hyperparameters were set to the default CellViT configuration for training on the PanNuke dataset \cite{horst2024cellvit}. Specifically, we used the same optimizer, learning rate schedule, input patch size, loss functions, and number of training epochs as described in the official implementation for PanNuke. 
The batch size was set to 256, as the number of patches in our dataset is larger than in PanNuke.
We set the clustering resolution to $r = 4.0$. For gene voting, the top-$N = 10$ differentially expressed genes (DEGs) were used for classification with rank-based weighting enabled. For the neoplastic check, the top-$M = 20$ DEGs were used, with thresholds of $\tau_\text{vote} = 5$ and $\tau_\text{cancer} = 0.25$.

\begin{table}[t]
\centering
\caption{Dataset statics. *: patches were randomly sampled from more than 20,000 whole-slide images (WSIs) for PanNuke.}
\label{tab:statics}
\begin{tabular}{lccccc}
\toprule
Dataset & Split & Num. of tissues & Num. of slide &Patches & Nuclei \\ 
\midrule
HEST1K (ST-based) & Train & 9 & 51& 373,214  & 6,643,746  \\
PanNuke (Manual) & Validation &19& *200,000& 7,901 & 196,468  \\ \hdashline 
Seg.-Eval. (HEST1K) & Test & 4 & 8 & 73,091 & 3,704,652 \\ 
Cls.-Eval. & Test &1 & 1 & 10,238 & 219,342 \\ 
\bottomrule
\end{tabular}
\end{table}

\noindent
{\bf Dataset.}
Table~\ref{tab:statics} summarizes the dataset statistics: HEST1K \cite{jaume2024hest} for training, PanNuke \cite{gamper2019pannuke}, Seg.-Eval, and Cls.-Eval \cite{janesick2025biomarker,janesick2023high}.
Model training was conducted on HEST1K (ST-based), which provides whole-slide H\&E images, DAPI-derived nuclear masks, and cell-level gene expression profiles obtained via Xenium. Validation was performed on the PanNuke dataset, which contains expert-curated nucleus instance segmentation masks and annotations for five cell types (Neoplastic, Inflammatory, Connective, Dead, and Epithelial).
To ensure a fair comparison, experiments were restricted to the tissues shared by HEST1K and PanNuke (Breast, Colon, Kidney, Liver, Lung, Ovarian, Pancreatic, Prostate, Skin, and Stomach), and the same PanNuke validation split was used for all methods.

\noindent
\textbf{Segmentation Evaluation (Seg.-Eval.).}
Segmentation generalization was evaluated on four organs from HEST1K 
that are not included in Training dataset.

\noindent
\textbf{Classification Evaluation (Cls.-Eval.).}
For classification evaluation, we used a manually annotated 
breast cancer Xenium dataset \cite{janesick2025biomarker,janesick2023high}, which also includes whole-slide images, masks, and gene expression data.
Since the manual annotation included 14 fine-grained labels, we mapped them to four cell types based on cell type: Neoplastic, Inflammatory, Connective, and Epithelial.
As there was no label corresponding to dead cells, the “Dead” category was excluded.

\begin{table}[t]
\centering
\caption{Segmentation evaluation on Seg.-Eval. comparing supervised (trained on PanNuke dataset) and Ours.}
\label{tab:detection}
\begin{tabular}{lcccccccc}
\toprule
\textbf{Model} & \textbf{Dice} $\uparrow$ & \textbf{Jaccard} $\uparrow$ & \textbf{bPQ} $\uparrow$ & \textbf{F1} $\uparrow$ & \textbf{Prec.} $\uparrow$ & \textbf{Recall} $\uparrow$ \\
\midrule
CellViT (PanNuke)       & 0.6350 & 0.5220 & \textbf{0.4148} & 0.7492 & \textbf{0.7155} & 0.7862 \\
CellViT (HEST1K, Ours)  & \textbf{0.6585} & \textbf{0.5510} & 0.4018 & \textbf{0.7562} & 0.7028 & \textbf{0.8183} \\
\bottomrule
\end{tabular}
\end{table}

\noindent
{\bf Evaluation of Segmentation Performance.}
We evaluated the segmentation performance on unseen organs by comparing a model trained on the PanNuke dataset \cite{gamper2019pannuke} with our model trained using spatial transcriptomics data. For a rigorous comparison, both models utilized the identical CellViT architecture, with the PanNuke baseline trained according to official data splits. Validation was conducted on the corresponding folds as detailed in the Dataset section.

Tables \ref{tab:statics} and \ref{tab:detection} summarize the dataset statistics and segmentation results, respectively. While the PanNuke baseline was fully supervised across 19 organs, the proposed HEST1K model was trained on 9 organs. Despite the smaller number of slides and organs, HEST1K leverages substantially more nuclear instances per slide via DAPI-derived masks, thereby facilitating dense supervision.

As demonstrated in Table \ref{tab:detection}, the proposed method achieves performance comparable to the PanNuke baseline. Notably, the most significant improvement is observed in Recall, which increased from 0.7862 to 0.8183 (+0.0321). This suggests enhanced robustness in detecting true nuclei within unseen organs.

Furthermore, DAPI-derived nuclear masks from Xenium data provide dense, spatially aligned supervision that improves detection accuracy. Given that these annotations are automatically generated, the framework is inherently scalable and stands to benefit from the incorporation of additional data.
\begin{figure}[t]
    \centering
    \includegraphics[width=0.9\linewidth]{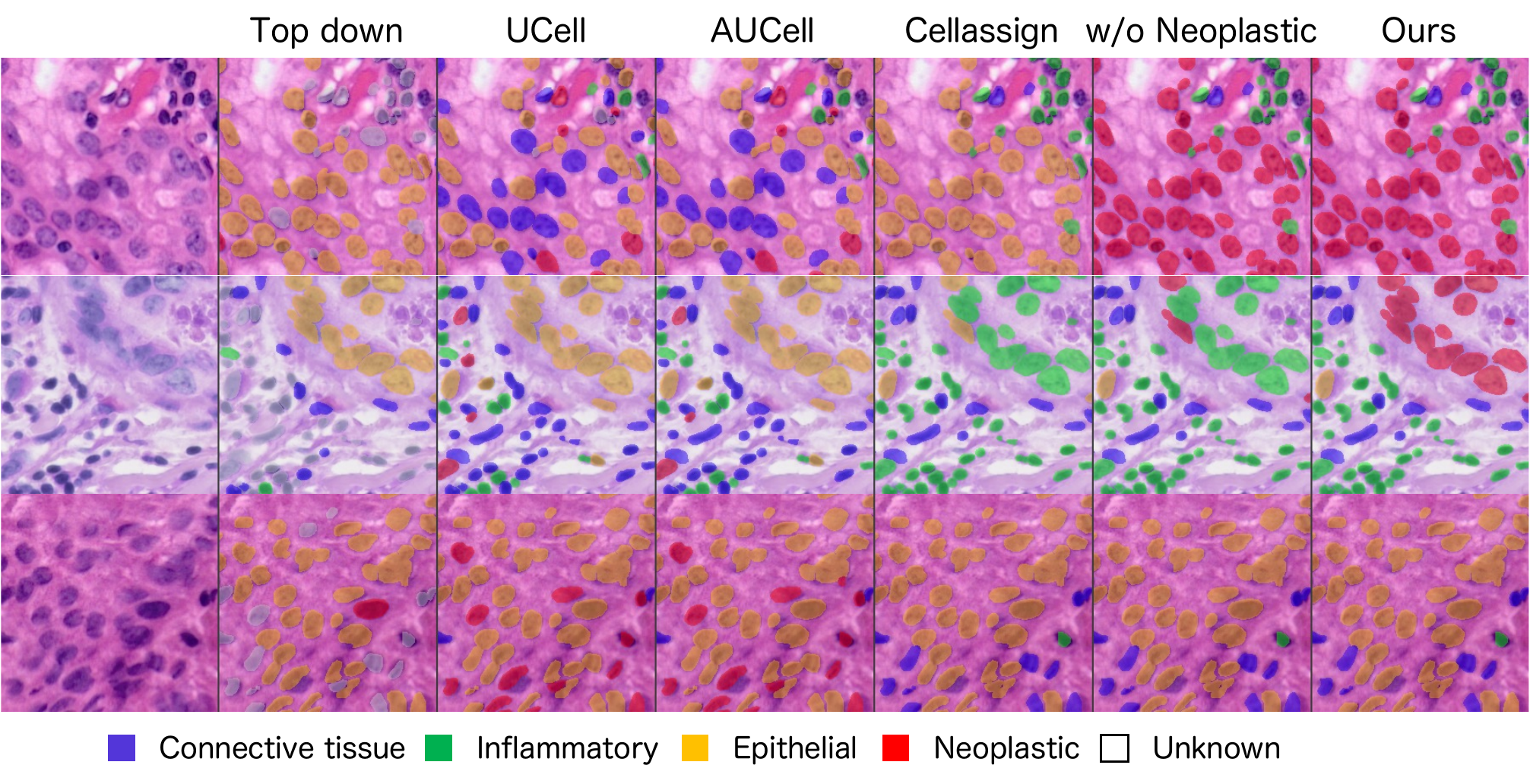}    
    \caption{Example of generated masks.}
    \label{fig:label_quality}
\end{figure}

\begin{figure}[t]
\centering

\begin{adjustbox}{valign=t}
\begin{minipage}{0.43\textwidth}
\centering
\captionof{table}{Classification performance for each model}
\label{tab:cls_comparisons}

\begin{tabular}{lcc}
\toprule
\textbf{Model} & \textbf{Macro F1} & \textbf{Acc} \\
\midrule
TopDown    & 0.1459 & 0.1261 \\
UCell      & 0.1297 & 0.1595 \\
AUCell     & 0.2309 & 0.2325 \\
CellAssign & 0.1919 & 0.2560 \\
\hdashline
w/o Hier.  & 0.3034 & 0.3041 \\
Ours       & 0.3492 & 0.5359 \\
\hdashline
PanNuke    & \textbf{0.3569} & \textbf{0.5696} \\
\bottomrule
\end{tabular}
\end{minipage}
\end{adjustbox}
\hfill
\begin{adjustbox}{valign=t}
\begin{minipage}{0.53\textwidth}
\centering
\includegraphics[width=\linewidth]{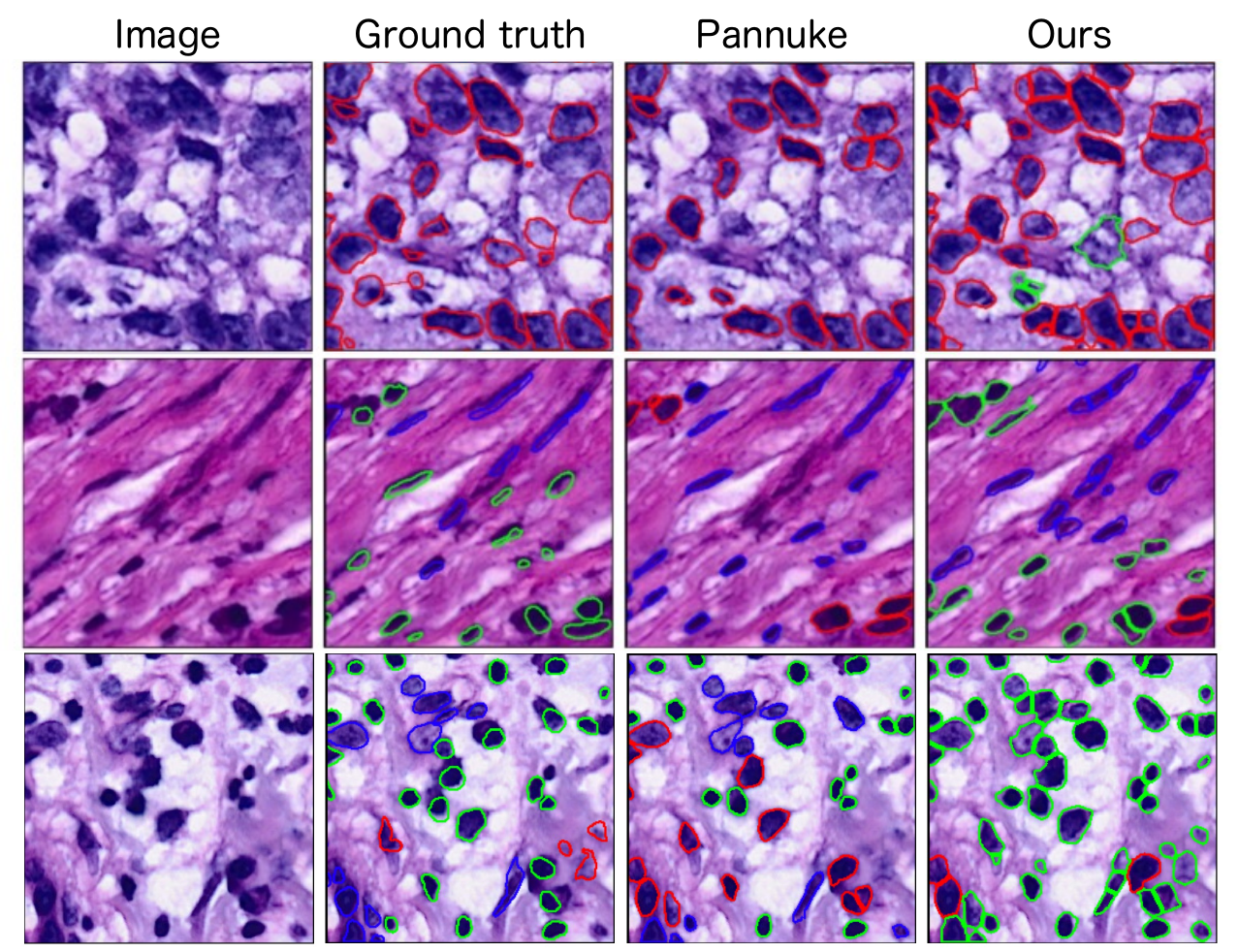}
\captionof{figure}{Example segmentation results.}
\label{fig:example}
\end{minipage}
\end{adjustbox}

\end{figure}

\noindent
{\bf Evaluation of Classification Performance.}
For cell-type classification, we compared our method with four reference-free approaches. Reference-based methods were excluded because they require organ-specific scRNA-seq datasets, whose availability varies across tissues.
We evaluated:
(1) a top-down approach implemented \cite{tirosh2016dissecting}, which assigns cell identities based on the average expression of predefined gene sets;
(2) UCell \cite{andreatta2021ucell}, a rank-based gene signature enrichment method;
(3) AUCell \cite{aibar2017scenic}, which computes enrichment using the area under the recovery curve; and
(4) CellAssign \cite{zhang2019probabilistic}, a Bayesian model that assigns cell types based on marker genes.
The same marker gene databases were used across all methods. As an ablation study, we also evaluated a variant without hierarchical classification of neoplastic cells (w/o Hier).
To compare with conventional supervised approaches, we additionally conducted experiments on the PanNuke dataset following the standard segmentation setup.

Fig. \ref{fig:label_quality} shows examples of labels generated by comparisons.
As shown in the top two rows of the results, the conventional method misclassifies neoplastic cells as epithelial. In contrast, the proposed method correctly classifies neoplastic cells through the proposed two-stage refinements. 

Table~\ref{tab:cls_comparisons} reports classification accuracy and macro F1 scores. The proposed method outperforms the reference-free baselines and improves upon the w/o Hier variant, demonstrating the effectiveness of hierarchical refinement. Performance remains slightly below the PanNuke-trained model, likely due to noise in gene expression–based automatic labels. Incorporating manually annotated datasets may further improve robustness.

Fig.~\ref{fig:example} presents qualitative comparisons between the PanNuke-trained model and the proposed method. While their classification tendencies differ, the proposed model correctly identifies some nuclei misclassified by the PanNuke-trained model.It indicate the feasibility of assigning classification labels that can support and enhance manual supervision in clinical image analysis workflows.

\section{Conclusion}
In this paper, we proposed a novel problem formulation for nuclei segmentation and classification in digital pathology that leverages spatial transcriptomics (ST) as an alternative to large-scale manual annotation.
By constructing a practical framework, we trained segmentation and classification models without conventional pixel-level annotations.
Experimental validation on a public dataset demonstrated the effectiveness of the proposed approach. The results indicate that ST-derived supervision can help for scalable training.

\noindent
{\bf Acknowledgements:} This work was supported by JSPS KAKEN JP24KJ2205, 
23H03862, and JP23K16991, JST ASPIRE JPMJAP2403, AMED grant \\ JP24ama221609h0001(PPROMOTE) (to YK) and JP25gm7010014h0001(PRIME), National Cancer Center Research and Development Fund 2024-A-6 (to YK).

\bibliographystyle{splncs04}
\bibliography{myrefs}

\end{document}